\documentclass{article} %
\usepackage{iclr2026_conference,times}
\iclrfinalcopy
\usepackage{amsmath,amsfonts,bm}

\def\eqref#1{equation~\ref{#1}}

\def\1{\bm{1}}

\DeclareMathAlphabet{\mathsfit}{\encodingdefault}{\sfdefault}{m}{sl}
\SetMathAlphabet{\mathsfit}{bold}{\encodingdefault}{\sfdefault}{bx}{n}

\usepackage{hyperref}
\usepackage{url}
\usepackage{fix-cm} %

\usepackage{import}
\usepackage{macros}
\usepackage{amsmath}
\usepackage{tikz}
\usepackage{amssymb} 
\usetikzlibrary{arrows.meta,shapes,positioning,quotes}
\usepackage{amsthm}

\newtheorem{definition}{Definition}
\newtheorem{property}{Property}
\newtheorem{problem}{Problem}
\newtheorem{theorem}{Theorem}
\newtheorem{corollary}{Corollary}

\usepackage{fontawesome5}
\usepackage[most]{tcolorbox}
\usepackage{multirow}
\usepackage{soul}
\usepackage[linesnumbered,ruled,vlined]{algorithm2e}
\usepackage{wrapfig}
\usepackage{enumitem}

\title{LLM Chemistry Estimation for Multi-LLM Recommendation}

\author{Huascar Sanchez, Briland Hitaj \\
Computer Science Laboratory\\ 
SRI International\\ 
Menlo Park, CA 94025 USA\\
\texttt{\{huascar.sanchez, briland.hitaj\}@sri.com}
}

\begin{document}

\maketitle

\begin{abstract}
Multi-LLM collaboration promises 
accurate, robust, and context-aware
solutions, 
yet existing approaches rely on implicit
selection and output assessment without analyzing whether collaborating models truly
complement or conflict. 
We introduce \emph{LLM Chemistry} -- 
a framework 
that measures
when LLM combinations exhibit synergistic or
antagonistic behaviors 
that shape collective performance beyond individual
capabilities. 
We formalize the notion of chemistry among LLMs, propose
algorithms that quantify it by analyzing interaction dependencies,
and recommend optimal model ensembles accordingly. 
Our theoretical analysis shows
that chemistry among collaborating LLMs is most evident under heterogeneous model profiles, with its outcome impact shaped by task type, group size, and complexity. 
Evaluation on classification, summarization, and program repair tasks provides initial evidence 
for these task-dependent effects,
thereby reinforcing our theoretical results.
This establishes LLM Chemistry
as both a diagnostic factor in
multi-LLM systems and 
a foundation for ensemble recommendation.
\end{abstract}

\section{Introduction}
\label{sec:introduction}

Large Language Models (LLMs) are increasingly capable across tasks from code
generation to open-domain question answering. Yet no single model excels
universally~\citep{chang2024survey}. Different LLMs bring varied
strengths---reasoning, generation, or domain expertise---spurring interest in
multi-LLM collaboration: systems that coordinate multiple models to solve tasks
collectively. By combining complementary abilities, such collaboration promises
more accurate, robust, and context-aware solutions than any single model
alone~\citep{feng2024don,tran2025multi}, though 
a key challenge lies in
selecting ensembles that reliably exploit these strengths.

To address this challenge, 
a growing body of work has proposed numerous collaborative strategies positioned
at different stages of the inference pipeline: (1) before inference, 
\eg LLM routers~\citep{rosenbaum2017routing}; (2) during inference,
\eg ensemble decoding~\citep{li2024purifying,mavromatis2024pack}; and (3)
after inference, 
\eg LLM Cascades~\citep{chen2023frugalgpt,yue2023large} and LLM compensatory
cooperation~\citep{zhao2024explainability}. These strategies aim to harness the
strengths of multiple and frequently collaborating LLMs.

Existing methods focus primarily on selecting \emph{strong individual
models}---often large, high-performing closed-source LLMs such as ChatGPT,
Claude, and Gemini---
without considering
how their interactions affect group
performance~\citep{hu2024routerbench}.
Yet evidence shows 
these interactions can substantially influence
collaborative outcomes~\citep{liu2024dynamic} and even transmit unrelated
behavioral traits 
via
generated data~\citep{cloud2025subliminal}. Work on
compound AI systems has examined LLM assignment across modules under resource
constraints, but 
assumes modular independence and ignores potential
synergy---or interference---when models collaborate on a single task.
To illustrate how such dynamics play out in practice, consider a \textit{statement credibility classification} task:
``\textit{Small businesses create 70 percent of the jobs in America.} (Eric
Cantor).'' A single model may be precise but brittle (e.g.,
misclassifying unusual variations of this claim as \textit{false}), while
another may be broader but noisier (e.g., giving inconsistent answers such as
\textit{true, mostly true, half true}). Individually, neither suffices. Yet when
models are combined and allowed to \textit{interact through multiple rounds of
response generation and critique}, they may converge on a more accurate
answer---or entrench errors---showing that while multi-model solutions can
outperform individual models, their collective performance depends as much on
interaction as on individual accuracy~\citep{du2023improving}.

Motivated by this, we shift focus from allocation to interaction:
rather than treating LLMs as isolated units, we study how they perform jointly
on a single task, aiming to identify subsets that exhibit strong collaborative
synergy---what we call \textit{LLM Chemistry}. This raises a central question:
\textbf{\emph{For a given task, how can we identify which LLMs work best together, i.e., exhibit strong chemistry?}}
To address this question, we 
propose a framework for quantifying LLM Chemistry among LLMs collaborating on shared tasks. 
We formalize this notion and present algorithms that identify interaction dependencies among models and recommend those with strong chemistry.
Informally, an LLM $a$ interacts with an LLM $b$ if the benefit of using $a$ on
a task changes in the presence of $b$, and vice versa. We evaluate such chemistry purely through \textit{performance evaluation}, i.e., how effectively
LLMs achieve objectives together.
Our experiments across three benchmarks provide initial evidence that interaction dynamics shape collective performance---sometimes amplifying, sometimes constraining. This suggests model selection alone may be insufficient, since outcomes also depend on interaction effects as well as individual strength.
We model collaboration as a two-stage process of response generation and evaluation,
a common setup in
multi-LLM collaboration frameworks~\citep{du2023improving,madaan2023self,zhang2024chain}. In all benchmarks, LLMs interact through this process rather than via independent subgoal assignment.

We summarize our technical contributions as follows:
\begin{itemize}%
    \item We introduce and formalize the notion of LLM Chemistry and present
    \textsc{ChemE}, an algorithm for computing
    the LLM chemistry within a group of LLMs.
    \item We introduce the concept of Model Interaction Graphs (MIGs) as a mechanism to encode the performance and cost tradeoffs of different LLM interactions in \textsc{ChemE}.
    \item We present \textsc{Recommend}, an algorithm for selecting optimal LLM
    combinations that exhibit strong chemistry for collaborative tasks.
    \item We conduct empirical experiments across three diverse benchmarks
    demonstrating that interactions significantly influence collaborative
    performance and thus LLM Chemistry.
\end{itemize}

\section{LLM Chemistry}%
\label{sec:chem}

\subsection{Preliminaries}%
\label{subsec:prelimns}

\textbf{Basic Concepts}. Let $Q$ be a query that needs answering. Suppose we
have a group of candidate LLMs (denoted by $S$), each with recorded past
performance (e.g., accuracy of LLMs and a quality score for their outputs). Our
goals are: (1) to quantify how LLMs interact---their \textit{chemistry}---when
jointly answering $Q$; (2) to assess the impact of their interactions on
performance; and (3) to identify the optimal \textit{configuration} (i.e., subset of
LLMs) for $Q$. \fref{fig:overview-chemistry} illustrates this process.

Given a configuration $X \subseteq S$ chosen to answer $Q$, we introduce
$\mathit{cost}_Q(X)$ as the total cost incurred by using $X$, and
$\mathit{used}(X) \subseteq X$ as the subset of LLMs that successfully generated
an answer for $Q$. Note that $X \setminus \mathit{used}(X)$ may be non-empty if
some LLMs in $X$ did not contribute an output. The cost is defined as follows: assume $X$
produces $n > 0$ outputs for $Q$. Each output $o_i$ has a quality score $q_i \in
[0,10]$ (the consensus grade), each producing LLM has accuracy $a_i \in [0,1]$
(its reliability score; see below for how $q_i$ and $a_i$ are computed), and a
weight $w_i = 1/i$.\footnote{Weights are positive reals, inversely proportional
to rank, so higher-ranked answers add more to the cost.} Since $q_i$ is in
$[0,10]$, we normalize it to $[0,1]$ to match accuracy $a_i$:
$q_i^{\mathit{norm}} = q_i/10$. 
Accordingly, we define the per-answer \textit{penalty} (a measure of the joint error in quality and accuracy, which decreases as either improves) as $\mathit{penalty}_i = (1 - q_i^{\mathit{norm}})(1 - a_i)$.
The total cost over all $n$ answers,
weighted by $w_i$, is:

\begin{equation} %
	\label{eq:cost}
	\mathit{cost}_Q(X) = \sum_{i=1}^{n} w_i \cdot \mathit{penalty}_i \,
\end{equation}

The best-case scenario ($q_i^{\mathit{norm}} = 1, a_i = 1$): penalty is $0$. Lower values in
either quality or accuracy, increase the penalty (thus, the total cost). E.g.,
consider three answers with (rank, quality, accuracy): (1, 9.0, 0.90), (2, 8.0,
0.80), (3, 7.0, 0.70). Using weights $w_i = 1/i$, 
the total cost is $0.01 + 0.02 + 0.0297 = 0.0597$.
This shows low penalties for high-quality, high-accuracy answers, with penalties rising down the ranks; this cost analysis sets the stage for defining LLM benefit within an ensemble.

\textbf{Presence Benefit of an LLM in an Ensemble}. Given Equation~\ref{eq:cost}
and disjoint sets $X, Y \subseteq S$, the \textit{presence benefit} of an LLM is
the change in performance cost when $Y$ is selected in addition to $X$,

\begin{definition}[Benefit]
Given two LLM sets $X$, $Y \subseteq S$ and a query $Q$, the benefit of $X$
with respect to $Y$ and $Q$ is defined as $\mathit{benefit}_Q(X,\,Y) =
\mathit{cost}_Q(Y) - \mathit{cost}_Q(X\;\cup\;Y)$.
\end{definition}

The value of $\mathit{benefit}_Q(X, Y)$ can be negative if adding $X$ to $Y$
raises the total cost, indicating degraded rather than improved performance.
Understanding such degradations---and, more broadly, how LLMs affect each
other's effectiveness---is key to analyzing model chemistry.

\textbf{Quality and Accuracy Scores}. We define the quality of an output $o_i$ (its consensus score $q_i$) as the 
aggregated estimate of its quality from multiple LLMs, and the accuracy of an 
LLM ($a_i$) as how closely its evaluations align with these consensus grades. We compute quality scores $q_i$ using a
consensus-based approach inspired by the \textit{Vancouver crowdsourcing
algorithm}~\citep{de2014crowdgrader}. We apply the \textit{Minimum Variance
Linear Estimator} (MVLE) to iteratively infer consensus scores for each output
and a consensus variance for each LLM. The consensus score becomes $q_i$, with
higher values assigned to models whose outputs consistently align with
low-variance consensus. In this setup, as in MVLE, the \textit{inverse variance} serves as a
proxy for \textit{review accuracy}.
\textit{Generation accuracy} measures how well an LLM's outputs match ground truth reference answers, computed by direct comparison when available (values in $[0.0, 1.0]$), or set to $0.0$ when none exist. 

To complement this, we compute a \textit{continuous} accuracy $a_i$ for an LLM
by combining generation and review accuracy. With ground truth, we weight them
$75\%/25\%$; without it, $25\%/75\%$. For example, o3-mini on statement
classification has generation accuracy $1.0$ and review accuracy $0.846$,
yielding $a_i = 0.962$.
We use MVLE over traditional voting to compute $q_i$ and $a_i$ for its statistical 
grounding, robustness to noise, and ability to refine reliability estimates. 

\subsection{The LLM Chemistry Problem}%
\label{subsec:llmcp}

Beyond assessing individual contributions, we aim to understand how LLMs in a
set $S$ interact when jointly answering a query $Q$. These
interactions---whether cooperative or redundant---can significantly impact the
overall effectiveness of the ensemble. Intuitively, two LLMs $a$ and $b$
interact when their individual benefits are dependent. This can occur in two
ways. First, if $a$ and $b$ exhibit overlapping performance ($q_i$, $a_i$) profiles---i.e., they produce
similar or redundant outputs on $Q$---then one can substitute for the other.
Building on our notion of LLM chemistry, this constitutes a \emph{negative
chemistry}, as using both LLMs adds little benefit over using just one. Second,
if $a$ and $b$ contribute complementary outputs---e.g., one extracts factual
content while the other interprets it---their combination can yield a better
answer than either alone. This constitutes a \emph{positive chemistry}.

We formalize this intuition by defining the \textit{LLM Chemistry} between two
LLMs $a$ and $b$, denoted $\mathit{chem}_Q(a,b,S)$, as the change in the benefit
of $a$ when $b$ is added to a set $X$:

\begin{definition}[LLM Chemistry]%
\label{dfn:doc}
Given LLMs $a$, $b$ in $S$, and $X \subseteq S$ be a set of LLMs in $S$ such
that $X \cap \{a,\!b\} = \emptyset$, the chemistry between LLMs $a$ and $b$
\wrt $Q$ is defined as:

\begin{equation}
\label{eq:doc}
\mathit{chem}_Q(a,b,S) = \max_{X\subseteq S\setminus\{a,\!b\}} 
    \frac{\Delta(a,b,X)}{\mathit{cost}_Q(X \cup \{a,\!b\})}
\end{equation}

\noindent
where $\Delta(a,b,X)$ is the absolute difference between
$\mathit{benefit}_Q(\{a\}, X)$ and $\mathit{benefit}_Q(\{a\}, X \cup \{b\})$.
\end{definition}

This definition reflects how strongly LLMs $a$ and $b$ complement each other
when answering a query $Q$. Specifically, the LLM chemistry is highest when $a$
and $b$ provide highly complementary information, and lowest when their outputs
are redundant, conflicting, or erroneous. To ensure that $\mathit{chem}_Q(a,\!b,S)$
is independent from any specific choice of $X$, we define it explicitly as the
maximum interaction effect across all possible subsets $X \subseteq S \setminus
\{a,\!b\}$. This approach captures the worst-case interaction scenario,
ensuring the measure reflects the strongest potential chemistry between
two LLMs, independent of particular subset selections. For clarity, we omit $Q$ in $\mathit{cost}(X)$, $\mathit{benefit}(X,Y)$, and $\mathit{chem}(a,b,S)$ when it is clear from context.

The LLM chemistry between any pair of LLMs is critical to understanding how the
set $S$ influences answers to $Q$. This leads to a key problem addressed in this
paper:

\begin{problem}[LLMs Chemistry Problem (\textbf{LLMCP})]
Given a set of LLMs $S$, a subset $X \subseteq S$, and a threshold $\tau \geq
0.0$, find all pairs of distinct LLMs $a,\!b \in S$ such that
$\mathit{chem}(a,b,S) > \tau$ \wrt $Q$.
\end{problem}

In this problem, we aim to identify pairs of LLMs that exhibit sufficiently
strong chemistry for answering $Q$. The goal is not to compute exact chemistry
values, but to decide whether chemistry is sufficiently high. Detecting strong
chemistry requires finding only a single LLM configuration where threshold
$\tau$ is exceeded, while ruling it out requires checking all relevant
configurations. Thus, worst-case complexity remains as high as computing full
LLM Chemistry.

\section{Criteria for Optimal LLM Selection}%
\label{sec:optimal}

In our formulation of \textit{LLMCP}, LLM interactions are evaluated by $\mathit{cost}(X)$ (\eref{eq:cost}) to determine the optimal LLM selection for a query $Q$. This requires computing $\mathit{chem}(a,b,S)$ for every LLM pair $a,b$ across all subsets $X \subseteq S$, giving a lower bound of $\Omega(2^{|S|},|S|^{2})$.

In practice, the optimization of LLM selection should not depend on a particular
set $X$ in an arbitrary way. For instance, adding more LLMs to the available set
should not increase the cost of answering $Q$, since a larger set only broadens
the space of possible responses. To avoid such anomalies, we make a natural
assumption on $S$. Conceptually, we assume $S$ is made of diverse and
independently strong LLMs whose strengths compensate for one another's
weaknesses, rather than reinforce them. Informally, we say $S$ is
\textbf{diverse} if its members exhibit heterogeneous performance
characteristics---i.e., they differ meaningfully in their performance ($q_i$, $a_i$)
profiles. Conversely, $S$ is \textbf{homogeneous} if all models produce similar
responses with comparable accuracy (i.e., aligned performance profiles).

This diversity assumption has profound implications for cost function
behavior and thus on LLM Chemistry estimation. Under this assumption,
$\mathit{cost}(X)$ should exhibit structure that reflects the reliability and
diversity of the underlying models. Rather than treating all LLM combinations as equally unpredictable, we posit that cost varies predictably with model quality, accuracy, and set composition. This motivates
three key properties of the cost function. These properties---monotonicity,
linearity, and submodularity---not only align with practical expectations but
also play a central role in making the selection of optimal LLM sets
computationally feasible.

\subsection{Properties of the Cost Function}%
\label{subsec:properties}
We formalize three properties that characterize \eref{eq:cost}:

\begin{property}[Monotonicity]%
\label{property:cost_mon}
The cost of a set $X$ for answering $Q$, $\mathit{cost}(X)$, is
monotonically decreasing with the output quality scores $q_i$ and the accuracy values $a_i$ of the LLMs producing them.

\end{property}

This property means the cost decreases as an LLM's accuracy or output
quality increases, reflecting the intuition that accurate, high-quality outputs
lower the total cost. (Proof in Appendix~\ref{app:cost_proofs}.)

\begin{property}[Linearity of Cost]%
\label{property:cost_linearity}
The $\mathit{cost}(X)$ separates as a sum of terms, each depending
only on the quality $q_i$ of an individual LLM output and the accuracy $a_i$ of its producing LLM.
\end{property}

This property means the cost is additive: each LLM affects the total cost only
through its own contribution, with no cross-terms or interactions between
outputs. (Proof in Appendix~\ref{app:cost_proofs}.)

\begin{property}[Submodularity]
\label{property:submodularity}
For all sets $X \subseteq Y \subseteq S$ and any LLM $a \in S \setminus Y$,
\eref{eq:cost} satisfies: $\mathit{cost}(X) - \mathit{cost}(X
\cup \{a\}) \geq \mathit{cost}(Y) - \mathit{cost}(Y \cup \{a\})$.
\end{property}

This property implies diminishing returns: for any given LLM $a$, its marginal benefit is larger when added to a smaller set $X$ than to a larger superset $Y$. (Proof in Appendix~\ref{app:cost_proofs}.)

\subsection{Model Interaction Graphs (MIGs)}%
\label{subsec:migs}

\begin{wrapfigure}[14]{r}{.49\textwidth}
\vspace{-0.2cm}
	\centering
	\begin{tikzpicture}[
			scale=0.75, %
			node distance=2cm,
			every node/.style={
				draw,
				rounded corners,
				fill=gray!10,
				inner sep=2pt,
				minimum height=0.1cm,
				font=\small\ttfamily  %
			},
			solid edge/.style={-Stealth, thick},
			dashed edge/.style={-Stealth, thick, dashed}
		]

		\node (abc) at (0,0) {\underline{a},b,\underline{c}:0.05};
		\node (ab) at (-2.5,-2) {\underline{a},b:0.08};
		\node (ac) at (0,-2) {a,\underline{c}:0.07};
		\node (bc) at (2.5,-2) {\underline{b},c:0.006};
		\node (a) at (-2.5,-4) {a:0.010};
		\node (b) at (0,-4) {b:0.012};

		\draw[solid edge] (abc) -- (ab);
		\draw[solid edge] (abc) -- (ac);
		\draw[solid edge] (abc) -- (bc);

		\draw[solid edge] (ab) -- (a);
		\draw[solid edge] (ab) -- (b);

		\draw[solid edge] (ac) -- (a);
		\draw[solid edge] (bc) -- (b);
	\end{tikzpicture}
\caption{MIG for $S\!=\!\{a,b,c\}$. Underlined elements indicate $\mathit{used}(X)$ 
(LLMs with $a_i \geq 0.5$). Sample cost values are provided in each node.
}
        \label{fig:MIG}
\end{wrapfigure}
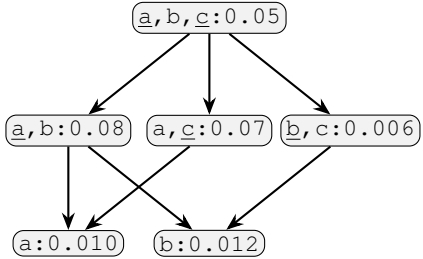

To efficiently compute LLM chemistry, we introduce the \textit{Model Interaction
Graph} (MIG), a directed acyclic graph (DAG) over subsets of $S$. The MIG
encoding draws inspiration from the \textit{Index Benefit Graph} (IBG)
\citep{frank1992adaptive}. \fref{fig:MIG} shows an MIG for $S\!=\!\{a,b,c\}$.

A central property of MIGs, inherited from IBGs, is that reasoning about subsets does not
require every subset to be explicitly represented: larger nodes contain enough
information to compute $\mathit{cost}(X)$ and $\mathit{used}(X)$ for overlapping
subsets, allowing the MIG to compactly encode all necessary information and
avoid exponential blowup. Each node in the MIG represents an LLM set $X
\subseteq S$ and stores both $\mathit{used}(X)$ and $\mathit{cost}(X)$. Edges capture how costs change as LLMs are added or
removed. Nodes and edges are constructed hierarchically, starting from the full
set $S$; at each step, for each node $X$ and used LLM $a \in \mathit{used}(X)$,
we create a new node $X' = X \setminus {a}$ with a directed edge from $X$ to
$X'$, unless $X'$ already exists.
\section{Computing LLM Chemistry for Multi-LLM Recommendation}%
\label{sec:approach}

\subsection{Computing LLM Chemistry}%
\label{subsec:doc-cover}

\begin{wrapfigure}[26]{r}{.55\textwidth}
\vspace{-1.5em}
\noindent
\begin{minipage}{.53\textwidth}
\hspace{5pt}
\begin{algorithm}[H]
\caption{Computes $\mathit{chem}(a,b,S)$ for all LLM pairs $a, b$ using memoized MIG lookups. $\mathcal{G}_Q$ encodes subset relationships. Solves \textbf{LLMCP}. Sets $\tau = 0$ to retain all chemistry information for later analysis.}
\label{fig:mdoc}
\fontsize{8.}{9}\selectfont
\DontPrintSemicolon
\SetKwInOut{Function}{Function}
\SetKwInOut{Input}{Input}
\SetKwInOut{Output}{Output}
\SetKwInOut{Data}{Data}

\Function{$\mathit{\textsc{ChemE}}$}
\Input{A set of LLMs $S$.}
\Output{The LLM chemistry for each distinct $a,\!b \in S$.}
\Data{Hash table $t_Q : S \times S \rightarrow \mathbb{R}$. Memoized node-and-cost lookup.}

Initialize $t_Q[a, b] \leftarrow 0$ for each distinct $a, b \subseteq S$\;
Construct the MIG $\mathcal{G}_Q$ for $Q$\;
\For{$k \leftarrow 0$ \KwTo $|S|$}{
  \ForEach{$X \subseteq S$ with $|X| = k$}{
    Let $Y \supseteq X$ in $\mathcal{G}_Q$ and $\mathit{cost}(Y)$ be memoized\;
    \If{$Y$ is undefined}{\textbf{continue}}
    \ForEach{distinct $a, b \in S - X$}{
      Let $X_a \leftarrow X \cup \{a\}$, $X_b \leftarrow X \cup \{b\}$, $X_{ab} \leftarrow X \cup \{a, b\}$\;
      Memoize $Y_a \supseteq X_a$, $Y_b \supseteq X_b$, $Y_{ab} \supseteq X_{ab}$ in $\mathcal{G}_Q$ and their costs\;
      \If{any of $Y_a$, $Y_b$, $Y_{ab}$ are undefined}{\textbf{continue}}
      $d \leftarrow \frac{|\mathit{benefit}(\{a\}, Y) - \mathit{benefit}(\{a\}, Y_b)|}{\mathit{cost}(Y_{ab})}$\;
      $t_Q[a, b] \leftarrow \max\{t_Q[a, b], d\}$\;
      $t_Q[b, a] \leftarrow \max\{t_Q[b, a], d\}$\;
    }
  }
}
\Return{$\mathit{chem}(a,b,S) = t_Q[a, b]$ for each $\{a, b\} \subseteq S$}\;

\end{algorithm}
\end{minipage}
\end{wrapfigure}
In this section, we introduce \textsc{ChemE} (\aref{fig:mdoc}).
Given a query $Q$,  
\textsc{ChemE} builds a MIG and looks at its structure
to compute $\mathit{chem}(a,b,S)$ for every pair of LLMs in $S$ \wrt $Q$.
It does so by iterating over
all subsets $X \subseteq S$ to compactly encode
costs associated with various combinations of LLMs over relevant subsets of $S$.
\textsc{ChemE} explicitly enumerates subsets to compute this
maximum value, directly aligning with \dref{dfn:doc}.

For example, given the MIG in \fref{fig:MIG}, assume additional hypothetical costs
$\mathit{cost}(\{c\}) = 0.15$ and $\mathit{cost}(\emptyset) = 0.20$, and the
case where $X = \{c\}$. Using the MIG costs, we have $\mathit{benefit}(\{a\},
\{c\})=0.08$, $\mathit{benefit}(\{a\}, \{b,c\})=0.01$, yielding
$\mathit{chem}(a,\!b,\{c\})=\frac{|0.08 - 0.01|}{0.05}=1.4$. 
As $\{c\}$ is the only relevant subset, the LLM chemistry for ($a$,
$b$) is also $1.4$.
\textsc{ChemE} yields the following main result:
\begin{theorem}
\label{thm:chem}
LLM Chemistry emerges in $S$ as a function of the MIG iff models
exhibit heterogeneous performance $(q_i,a_i)$ profiles; for (near-)identically performing models, cost-based selection 
pressure vanishes, so no interaction effects can be detected (chemistry = $0$).
\textnormal{(Proof in Appendix~\ref{sec:theory})}.
\end{theorem}

This novel, counter-intuitive finding suggests that model diversity
(i.e., model performance diversity) is essential for chemistry estimation:
near-maximal performance makes chemistry-based multi-LLM recommendation
unnecessary, while near-zero performance makes them infeasible.
A direct consequence of Theorem \ref{thm:chem} is that chemistry
scales with model performance diversity, as formalized in Corollary \ref{cor:chem}: 

\begin{corollary}
\label{cor:chem}
Chemistry in $S$ varies monotonically with model performance diversity:
as diversity increases, chemistry either always increases or always 
decreases, depending on task type. 
\textnormal{(Proof in Appendix~\ref{sec:theory})}
\end{corollary}

Theorem~\ref{thm:chem} and Corollary~\ref{cor:chem} establish the theoretical
basis for LLM Chemistry, which we now operationalize in the next subsection to
select effective ensembles for a given query, directly addressing the central
question posed in the introduction.

\subsection{Multi-LLM Recommendation}%
\label{subsec:recommend}

\begin{figure}[ht]
    \centering
    \includegraphics[width=.65\columnwidth]{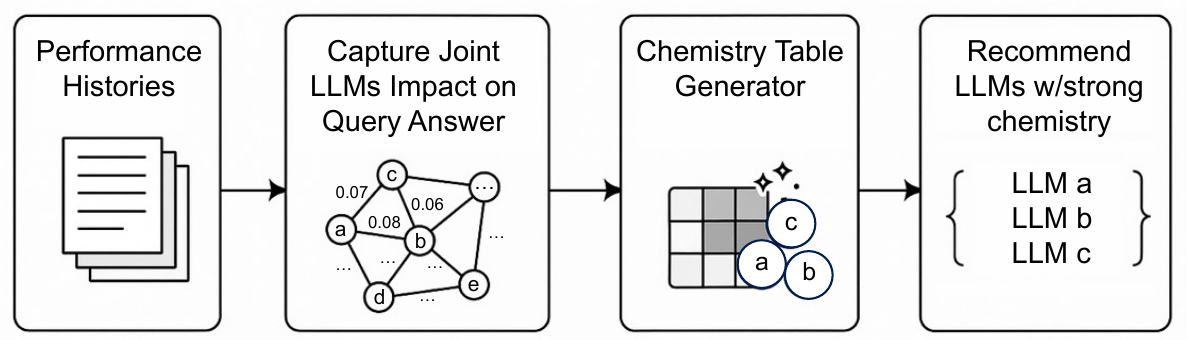}

    \caption{Illustration of LLM Chemistry estimation process for multi-LLM recommendation. A snapshot of the performance histories is provided in Appendix~\ref{ssec:trial-runs}.}
    \label{fig:overview-chemistry}
\end{figure}

We illustrate our LLM Chemistry estimation process for multi-LLM recommendation
process in Figure~\ref{fig:overview-chemistry}. The figure outlines the main stages of a multi-LLM recommendation session, from an initial MIG construction for $Q$ from past performance histories to LLM Chemistry estimation, to the recommendation of an optimal LLM configuration.
The central goal of this process 
is to recommend a set of LLMs that maximizes the
benefit of 
collaboration while minimizing the cost of answering $Q$.
To formalize this, 
we define
$\mathcal{X}=\{X_1, X_2, \ldots, X_n\}$ as a collection of non-empty subsets 
of $S$, where each $X_i \subseteq S$
is a candidate LLM configuration. 
\textit{Subsets may overlap, meaning the
same LLM can appear in multiple configurations}. 

To evaluate how well these subsets
group related LLMs, we define a loss function $\mathcal{L}(\mathcal{X})$
that measures unrealized chemistry potential for $\mathcal{X}$.
$\mathcal{L}(\mathcal{X})$ has two components:
(1)
inter-subset loss $\mathcal{L}_{inter}(\mathcal{X})$, 
the potential lost between LLMs placed into different subsets, 
and (2) intra-subset loss $\mathcal{L}_{intra}(\mathcal{X})$, 
the potential not realized within a subset compared to the theoretical maximum.
Two parameters control these trade-offs: $\alpha \in [0,1]$ balances the components, and $\beta > 0$ penalizes larger subsets to favor smaller, targeted recommendations (\eref{eq:error}).
\begin{equation}
\label{eq:error}
\mathcal{L}(\mathcal{X}) = \alpha \cdot \mathcal{L}_{inter}(\mathcal{X}) + (1-\alpha) \cdot \mathcal{L}_{intra}(\mathcal{X}) + \beta \cdot |\mathcal{X}|
\end{equation}
\vspace{-0.35cm}
\begin{align}
\label{eq:error-0}
\mathcal{L}_{inter}(\mathcal{X}) = \sum_{\substack{a,b \in S \\ a \neq b, \text{ different subsets}}}\mathit{chem}(a,\!b,S) - \sum_{\substack{X_i, X_j \in \mathcal{X} \\ i < j}} \sum_{a \in X_i} \sum_{\substack{b \in X_j \\ b \neq a}} \mathit{chem}(a,\!b,S)
\end{align}
\vspace{-0.25cm}
\begin{align}
\label{eq:error-1}
\mathcal{L}_{intra}(\mathcal{X}) = \sum_{a < b \in S} \mathit{chem}(a,\!b,S) - \sum_{X_i \in \mathcal{X}} \sum_{a < b \in X_i} \mathit{chem}(a,\!b,S)
\end{align}

These loss terms capture unrealized chemistry from separation (inter-subset) and incomplete groupings (intra-subset\footnote{In \eref{eq:error-1}, $a < b \in S$ denotes an arbitrary ordering, with each unordered pair ${a,b}$ counted once.}) of LLMs in $\mathcal{X}$. In this paper, we set $\alpha = 0.5$ to weight them equally and $\beta = 0.5$ to moderately penalize large subsets, typically recommending up to $10$ LLMs.

\aref{fig:recommend} recommends the optimal LLM subset from historically-derived
configurations by leveraging pre-computation and the error function in
\eref{eq:error}. Rather than exploring the exponential space of all possible
subsets ($2^{|S|}$), it evaluates only $n$ historical subsets as starting points
and performs local optimization through hill climbing with at most $iters$
iterations per starting point. 
This algorithm has a runtime complexity of $O(n
\times iters \times |S|^{3})$, which is polynomial in the input size and
exponentially better than the brute-force approach of evaluating all subsets.
This algorithm systematically explores the search space by starting from each
historical subset in $\mathcal{X}$, and applying local search through single LLM
additions, removals, or swaps to find a better configuration. It uses the
\textsc{loss} function to evaluate each candidate solution and guide the
optimization process. 
Specifically, \textsc{loss} computes intra-subset
chemistry as $\sum_{a < b \in X} \mathit{chem}(a,b,S)$ and inter-subset
chemistry as $\sum_{a \in X}\sum_{b \in S \setminus X}\mathit{chem}(a,b,S)$.
\begin{wrapfigure}[24]{r}{.5\textwidth}
\vspace{-1.3em}
\noindent
\begin{minipage}{.47\textwidth}
\hspace{7pt}
\begin{algorithm}[H]
\caption{Optimal LLM Subset Recommendation for $Q$.}
\label{fig:recommend}
\fontsize{8.}{9}\selectfont
\DontPrintSemicolon
\SetKwInOut{Function}{Function}
\SetKwInOut{Input}{Input}
\SetKwInOut{Output}{Output}
\Function{$\mathit{\textsc{Recommend}}$}
\Input{LLMs $S$, query $Q$, factor $\alpha$, max iters.}
\Output{Optimal subset $X^*$ for $Q$.}
$best\leftarrow \emptyset$; $minLoss\leftarrow \infty$\;
Get $\mathcal{X}_{Q} = \{X_1, \ldots, X_n\}$ from past runs\;
\tcp{Precompute using \textsc{ChemE}}
$maxT\leftarrow \sum_{a < b \in S} chem(a,b,S)$\;
$maxI\leftarrow \sum_{a,b \in S, a \neq b} chem(a,b,S)$\;
\ForEach{$X_i \in \mathcal{X}_{Q}$}{
    $cur\leftarrow X_i$; $loss\leftarrow \textsc{loss}(cur, maxT, maxI, \alpha)$\;
    \If{$loss < minLoss$}{
        $best \leftarrow cur$; $minLoss \leftarrow loss$\;
    }
    \tcp{Local search via hill climbing}
    \For{$iter \leftarrow 1$ \KwTo $iters$}{
        $bestN\leftarrow null$; $minN\leftarrow \infty$\;
        \ForEach{neighbor $X'$ of $cur$}{
            $lossN\leftarrow \textsc{loss}(X', maxT, maxI, \alpha)$\;
            \If{$lossN < minN$}{
                $bestN \leftarrow X'$; $minN \leftarrow lossN$\;
            }
        }
        \If{$minN \geq loss$}{\textbf{break}\;}
        $cur\leftarrow bestN$; $loss\leftarrow minN$\;
        \If{$loss < minLoss$}{
            $best \leftarrow cur$; $minLoss \leftarrow loss$\;
        }
    }
}
\Return{$best$}\;
\end{algorithm}
\end{minipage}
\end{wrapfigure}
These become loss terms by subtracting from the theoretical maxima ($maxT$ and $maxI$, using all LLMs in $S$), then combining per \eref{eq:error}.

In homogeneous cases---i.e., (near-)identically performing models---LLM
chemistry evaluates to zero by Theorem~\ref{thm:chem}. Thus the
\textsc{Recommend} algorithm either selects a single LLM configuration above the
threshold $\tau$ when all models are strong, or yields none if all are weak. 
\section{Experiments}
\subsection{Experiment Setup}%
\label{sec:exp_setup}

We evaluate our approach on three key aspects: (1) ensemble
effectiveness \breakdash whether chemistry-based selection outperforms baseline
selection strategies in their respective tasks, (2)
chemistry-complementarity correlation---how well chemistry scores predict
\textit{ensemble complementarity}\footnote{We define ensemble complementarity as
the extent to which LLM ensemble members contribute different strengths,
providing better coverage of accuracy-quality trade-offs than any single model.
We measure this using hypervolume (S-metric~\citep{zitzler2002multiobjective})
and Rao's quadratic entropy~\citep{rao1982diversity}, combined into a single
index balancing performance coverage with member diversity.}, and (3)
interaction effects across task complexity---how chemistry differs by task and
difficulty.
\textbf{Datasets:} We evaluate on three benchmark datasets across diverse
collaborative LLM scenarios: (1)~\textit{Liar benchmark}~\citep{wang2017liar}
for (English) statement credibility classification ($4,000$ statements with context and
labels); (2)~\textit{MTS-Dialog benchmark}~\citep{mts-dialog} for clinical notes
summarization ($1,700$ doctor-patient conversations and notes); and (3)~\textit{Quixbugs benchmark}~\citep{lin2017quixbugs} for 
automated program repair ($40$ buggy code snippets with fixes).

\textbf{Task complexity:} Datasets span different domains, and we
categorize their \textit{task complexity} by considering both computational
resources and reasoning effort. Following \citep{qi2025quantifying}, we
distinguish three levels of task complexities: low (minimal reasoning or compute), medium (extended
reasoning or compute), and high (substantial reasoning or compute, often beyond
model capability). Under this characterization, \textit{Liar} is low,
\textit{MTS-Dialog} medium, and \textit{Quixbugs} high.

\textbf{Models:} We evaluate on a diverse pool of LLMs spanning different model
families and capabilities: GPT-4o, o1/3/4-mini, Claude Sonnet 3.5/3.7, Gemini
2.0 flash (closed-source models), and Llama 3.1/3.3 70B, Mixtral 8x22B,
FireFunction v2, Qwen 2.5 32B (open-weights models). We selected these
models to capture variation in architectures, skills, and resource demands
across open- and closed-weight ecosystems, providing a representative basis for
evaluation.

\textbf{Metrics:} We report (1) Ensemble Effectiveness and Improvement over Best
Baseline - 
overall ensemble accuracy via soft voting over model accuracies (correct if
mean accuracy $>0.5$, majority rule), (2) Ensemble complementarity - extent of diverse strengths
among ensemble members, and (3) Chemistry correlation coefficients measuring
chemistry-complementarity relationships.

\textbf{Protocol:} For each benchmark, we evaluate all selection strategies
across three group size configurations: $N=3, 5, 10$ models. For each size
configuration, we conduct $10$ independent trials per selection strategy, with
each trial executing across $10$ records, resulting in $\approx 1,800$ total
experimental task executions across all configurations (exact number varies due
to likely execution failures). Selection strategies include: (1) Remote -
closed-source models, (2) Local - open-weights models, (3) Random - randomly
selected models, (4) Performance - top-k best individual performers, and (5) Chemistry
- models recommended by our approach. Performance metrics are aggregated across
tasks within each trial. Performance metrics are aggregated per trial, with
ANOVA used for strategy comparisons and Pearson correlations for
chemistry-complementarity, with significance at $p < 0.05$.
These baselines follow LLM ensemble practice~\citep{lu2024merge}:
random and top-k are standard, while remote vs. local reflects deployment
constraints often discussed in applied studies.

\reveal{
\todo{Optional: Things we might mention. 1. Hardware, Mem, and OS used. 2. Some anonymous URL to fake data/code of chemistry. 3. Mention that why we are not including the code for executing LLMs? Maybe we can mention we use Langchain to do this..}
}
\subsection{Results and Analysis}%
\label{sec:exp_results}

Table~\ref{tab:rq1_performance} evaluates whether chemistry-based
selection improves ensemble effectiveness compared to baseline strategies. For
statement credibility classification, chemistry yields measurable gains over
baselines ($+14.9\%$ at $N=5$, $+2.6\%$ at $N=10$), with the largest gains in
ensembles of size $5$. For automated program repair, chemistry matches baseline
effectiveness across group sizes ($0.0\%$), reflecting ceiling effects where
performance has saturated. For clinical note summarization, chemistry
underperforms in small groups ($-30.0\%$ at $N=3$) but converges to baseline in
larger ensembles ($+0.2\%$ at $N=5$, $-0.1\%$ at $N=10$). 
These results indicate
that chemistry's contribution to effectiveness is positive in some tasks and
ensemble sizes, neutral in others, and negative when ensembles are small or
saturated, underscoring its task- and size-dependent role in ensemble
effectiveness.

\begin{table}[htbp]
\centering
\caption{Comparison of Ensemble Effectiveness: Chemistry-Based vs. Baseline Selection (PERF.)}
\label{tab:rq1_performance}

\begin{tighttable}
\begin{tabular}{ccccc}
\hline
\textbf{Task} & \textbf{Group Size ($N$)} & \textbf{Chemistry} & \textbf{Best Baseline} & \textbf{Improvement ($\Delta$\%)} \\
\hline
\multirow{3}{*}{\shortstack{Statement Credibility \\ Classification}} 
 & $3$  & $0.547$ & $0.542$ & $+0.9\%$ \\
 & $5$  & $0.680$ & $0.592$ & $+14.9\%$ \\
 & $10$ & $0.679$ & $0.662$ & $+2.6\%$ \\
\hline
\multirow{3}{*}{\shortstack{Clinical Notes \\ Summarization}} 
 & $3$  & $0.700$ & $1.000$ & $-30.0\%$ \\
 & $5$  & $1.000$ & $0.998$ & $+0.2\%$ \\
 & $10$ & $0.999$ & $1.000$ & $-0.1\%$ \\
\hline
\multirow{3}{*}{\shortstack{Automated Program \\ Repair}} 
 & $3$  & $1.000$ & $1.000$ & $+0.0\%$ \\
 & $5$  & $1.000$ & $1.000$ & $+0.0\%$ \\
 & $10$ & $1.000$ & $1.000$ & $+0.0\%$ \\
\hline
\end{tabular}
\end{tighttable}
\begin{minipage}{1.\linewidth}
\scriptsize
\textbf{Note:} Ensemble effectiveness = Overall ensemble accuracy (via soft
voting). Values are average ensemble effectiveness across all configurations for
each group size, comparing chemistry-based selection with the best baseline.
\textbf{Improvement ($\Delta$\%)} = Relative difference: $(\text{Chemistry} -
\text{Baseline}) / \text{Baseline} \times 100$. Chemistry-based = Algorithm~\ref{fig:recommend}'s output. See Table~\ref{tab:best_ensembles} for model
compositions of best ensembles.
\end{minipage}
\end{table}
Table~\ref{tab:chemistry_results} examines how well chemistry scores predict
ensemble complementarity. For statement credibility classification, chemistry is
positively associated with complementarity ($r=0.319$, $p<0.001$), suggesting
that higher chemistry scores correspond to ensembles that cover the
accuracy-quality trade-off more effectively.\footnote{Due to space limitations, visualizations of these trade-offs are provided in Appendix~\ref{sec:interpretation}.} By contrast, automated program
repair ($r=-0.154$, $p<0.01$) and clinical note summarization ($r=-0.226$,
$p<0.001$) show negative correlations, indicating that higher chemistry scores
are linked to ensembles with reduced model performance diversity. These findings
provide evidence that chemistry is a significant predictor of complementarity
across all three tasks, but the direction of the relationship differs: in
classification it signals beneficial complementarity, whereas in program repair
and summarization it reflects reductions in performance diversity consistent
with saturation effects where little additional complementarity can be realized.
\begin{table}[htbp]
\centering
\caption{\label{tab:chemistry_results}Chemistry-Ensemble Complementarity relationships by Task Type}

\begin{tighttable}
    \addtolength{\tabcolsep}{-2pt}
\begin{tabular}{lccccc}
\hline
\textbf{Task Type} & \textbf{Correlation ($r$)} & \textbf{Effect Size} & \textbf{p-value} & \textbf{Sig.} & \textbf{n} \\
\hline
Statement Credibility Classification & $0.319$ & Moderate & $0.000$ & $***$ & $400$ \\
Clinical Notes Summarization & $-0.226$ & Small-to-Moderate & $0.000$ & $***$ & $394$ \\
Automated Program Repair & $-0.154$ & Small-to-Moderate & $0.008$ & $**$ & $291$ \\
\hline
\end{tabular}
\end{tighttable}

\begin{minipage}{1.0\linewidth}
\scriptsize
\textbf{Note:} $r$ = Pearson's correlation coefficient. Results shown for
strongest significant relationships per task type. Significance (Sig.) levels: $***$
$p<0.001$, $**$ $p<0.01$, $*$ $p<0.05$. \underline{\textit{Key finding:}}
LLM-Chemistry-ensemble complementarity relationship reverses by task type ($r =
-0.154$ vs $r = 0.319$).
\end{minipage}
\end{table}
We treat task complexity as an orthogonal lens to the task-type categories to examine whether chemistry effects scale with difficulty (Table~\ref{tab:rq3_complexity}).  
For statement credibility classification (low complexity),
chemistry scores correlate positively with complementarity as group size
increases ($r=0.118$, $p<0.05$ for $N = 5$; $r=0.319$, $p<0.001$ for $N = 10$),
while effectiveness shows no significant association. For
clinical note summarization (medium complexity), chemistry shows no association
in smaller groups but correlates positively with complementarity at $N=10$
($r=0.148$, $p<0.01$). For automated program
repair (high complexity), chemistry scores correlate negatively with
complementarity in medium groups ($r=-0.154$, $p<0.01$) but
positively with effectiveness across all group sizes ($N = 3$: $r=0.189$,
$p<0.001$; $N = 5$: $r=0.144$, $p<0.05$; $N = 10$: $r=0.229$, $p<0.001$). These results indicate that task complexity moderates
chemistry's role: in low-complexity tasks, chemistry enhances complementarity
without affecting effectiveness; in
medium-complexity tasks, it begins to matter only for larger ensembles; and in high-complexity tasks, chemistry reduces
complementarity but consistently improves effectiveness.
\begin{table}[htbp]
\centering
\caption{\label{tab:rq3_complexity}Task Complexity and Chemistry Effects on Ensemble Performance}

\begin{tighttable}
    \addtolength{\tabcolsep}{-2.5pt}
\begin{tabular}{ccccccc}
\hline
\textbf{Task (Complexity)} & \textbf{Group Size ($N$)} & \textbf{Complementarity ($r$)} & \textbf{Sig.} & \textbf{Effectiveness ($r$)} & \textbf{Sig.} & \textbf{n} \\
\hline
\multirow{3}{*}{\shortstack{Statement Credibility \\ Classification (Low)}} 
 & $3$ & $0.089$ & ns & $0.014$ & ns & $368$ \\
 & $5$ & $0.118$ & $*$ & $-0.068$ & ns & $358$ \\
 & $10$ & $0.319$ & $***$ & $-0.024$ & ns & $400$ \\
\hline
\multirow{3}{*}{\shortstack{Clinical Notes \\ Summarization (Medium)}} 
 & $3$ & $0.000$ & ns & $0.000$ & ns & $383$ \\
 & $5$ & $0.000$ & ns & $0.000$ & ns & $357$ \\
 & $10$ & $0.148$ & $**$ & $0.027$ & ns & $409$ \\
\hline
\multirow{3}{*}{\shortstack{Automated Program \\ Repair (High)}} 
 & $3$ & $-0.043$ & ns & $0.189$ & $***$ & $451$ \\
 & $5$ & $-0.154$ & $**$ & $0.144$ & $*$ & $291$ \\
 & $10$ & $-0.086$ & ns & $0.229$ & $***$ & $335$ \\
\hline
\end{tabular}
\end{tighttable}

\begin{minipage}{1.0\linewidth}
\scriptsize
\textbf{Note:} Values are Pearson's $r$ correlations between chemistry scores and
ensemble complementarity (Cols.~3-4) or effectiveness (Cols.~5-6). $n$~= number
of ensembles. Significance (Sig.): $***$ $p<0.001$, $**$ $p<0.01$, $*$ $p<0.05$, ns =
not significant.  
\underline{\textit{Key finding:}} Chemistry-ensemble correlations vary with task
complexity: complementarity dominates at low/medium complexity, effectiveness at
high complexity.
\end{minipage}
\end{table}
Together, the results from
Tables \ref{tab:rq1_performance}-\ref{tab:rq3_complexity} show that LLM
chemistry's contribution to ensembles is not uniform: (1) it improves effectiveness
in some tasks and sizes, (2) predicts complementarity in a task-dependent manner,
and (3) is further moderated by task complexity and group size.
\section{Related Work}%
\label{sec:related}

As multi-AI systems become increasingly prevalent, researchers have explored
\textbf{ensemble methods} for aggregating model
outputs~\citep{lu2024merge,
ICLR2024_327b9b8d,zhang2024chain,chen2025harnessing} and
solving various problems through model collaboration~\citep{miao2024specinfer,
zhou2024survey, feng2024don, zhao2024explainability, tran2025multi},
\textbf{multi-agent communication topologies} for improving operation efficiency
and effectiveness~\citep{li2023theory,ICLR2024_37c6d0bc,li2024improving,ICLR2024_578e65cd},
\textbf{automated selection techniques} for choosing optimal
models~\citep{rosenbaum2017routing,chen2023frugalgpt,yue2023large,dekoninck2024unified,ong2024routellm,patidar2025orchestration},
and \textbf{compound AI frameworks} for orchestrating models across pipeline
stages~\citep{chen2024more,santhanam2024alto,chen2025optimizing,chaudhry2025towards,wang2025aligning}.
However, these approaches typically assume model independence or modular
separation, overlooking the interaction dynamics that emerge when models
collaborate on shared tasks. Unlike these methods, our work explicitly models
the interaction effects between LLMs when they collaborate on shared tasks,
introducing the novel concept of LLM chemistry to capture both synergistic and
antagonistic relationships that influence collaborative performance.

\section{Conclusions and Future Work}
\label{sec:conclusions}

\reveal{\todo{Improve the conclusions text once all the experiments are complete!}}

We introduced the \emph{LLM Chemistry} estimation framework
for multi-LLM recommendations that adapts to task demands. We found
that chemistry can improve
ensemble effectiveness in some tasks and sizes, diminish under ceiling effects
where performance has saturated, and vary systematically with task complexity.
These structured, non-uniform outcomes are consistent with our theoretical
framework, showing that chemistry emerges under heterogeneous performance
profiles and manifests differently by task type, ensemble size, and complexity.
LLM chemistry estimation can be a
\emph{meta-learning signal} for effective ensemble formation, indicating when
chemistry is likely to help and when it is not. Looking forward, chemistry-aware
modeling may improve next-generation architectures, from Mixture-of-Experts
(MoE) and multi-agent systems to human-machine collaboration. 
Alongside this,
linking individual LLM skills to chemistry could explain ensemble performance beyond the current ``sum of parts'' view.

\reveal{\briland{Perhaps a discussion on extending LLM Chemistry to collaborating LLM Agents would be useful here.}}
%
\section*{Ethics Statement}
\label{sec:ethics}

Our work introduces the notion of LLM Chemistry, a framework for assessing LLM
teaming capabilities and recommending optimal ensembles for reliable
collaborative task completion. We demonstrate its effectiveness across
state-of-the-art models and well-established benchmarks covering tasks of
increasing difficulty.

By moving beyond the assumption that only large, high-performance closed-source
LLMs guarantee strong performance, our approach enables ensembles that combine
both open- and closed-weight models, thereby broadening accessibility and
promoting fairness in multi-LLM systems.

We also recognize that LLM ensembles increase computational and energy costs,
which must be weighed against their potential gains in robustness. By guiding
efficient ensemble formation, our framework can help mitigate these costs and
support sustainability. Finally, while our methods aim to improve the
reliability of outputs, they could be misused to create more persuasive but
deceptive systems; we discourage such applications and stress the importance of
transparency in deployment.


%
\section*{Acknowledgments}


We thank our SRI colleagues, Deborah Shands and Karsten Martiny, for their invaluable feedback and comments, which significantly improved the content and presentation of the paper.

This research is based upon work supported in part by the Advanced Research Projects Agency for Health (ARPA-H), Defense Logistics Agency (DLA) under Contract Number SP4701-23-C-0073. Any opinions, findings and conclusions or recommendations expressed in this material are those of the authors and do not necessarily reflect the views of Advanced Research Projects Agency for Health (ARPA-H), Defense Logistics Agency (DLA), or the United States Government. The U.S. Government is authorized to reproduce and distribute reprints for governmental purposes notwithstanding any copyright annotation therein.

\bibliography{main}
\bibliographystyle{iclr2026_conference}

\appendix
\section{Appendix}

\subsection{Proofs of Cost Function Properties}
\label{app:cost_proofs}

We provide detailed proofs for the three properties of the cost function presented in Section~\ref{subsec:properties}.

\paragraph*{Proof of Property~\ref{property:cost_mon} (Monotonicity)}

\begin{proof}
Assume weights $w_i \geq 0$, each term contributing to the cost
is of the form $w_i \cdot (1 - q_i^{\mathit{norm}})(1 - a_i)$. This expression
decreases as either the output quality $q_i^{\mathit{norm}}$ increases or the LLM
accuracy $a_i$ increases, with the other held fixed. Since the total cost is a
weighted sum of such terms it is monotonically decreasing in both
$q_i^{\mathit{norm}}$ and $a_i$.
\end{proof}

\paragraph*{Proof of Property~\ref{property:cost_linearity} (Linearity)}

\begin{proof}
By \eref{eq:cost}, the total cost is defined as:
$\mathit{cost}(X) = \sum{i=1}^{n} w_i \cdot (1 - q_i^{\mathit{norm}})
\cdot (1 - a_i)$.
Each term depends solely on the quality score $q_i$ of single output $o_i$ and
the accuracy $a_i$ of the LLM that produced it. There are no terms involving
combinations of multiple outputs or LLMs. Therefore, the total cost is a linear
sum of independent contributions, one per output-LLM pair, as claimed.
\end{proof}

\paragraph*{Proof of Property~\ref{property:submodularity} (Submodularity)}

\begin{proof}
The cost $\mathit{cost}(X)$ is a weighted sum of penalties from the outputs
produced by the LLMs in $X$. Adding a new LLM $a$ to $X$ can only improve or
maintain the best available outputs, thus $\mathit{cost}(X \cup \{a\})
\leq \mathit{cost}(X)$. Since $X \subseteq Y$, the set $Y$ already has
more LLMs and thus better or more redundant outputs. As a result, the marginal
benefit of adding $a$ to $Y$ is smaller than the benefit of adding $a$ to $X$.
Formally, 
\[
\mathit{cost}(X) - \mathit{cost}(X \cup \{a\}) \geq
\mathit{cost}(Y) - \mathit{cost}(Y \cup \{a\}).
\]
Thus, the cost function is submodular.
\end{proof}

\subsection{Theoretical Analysis of Chemistry Emergence}
\label{sec:theory}

Having established the algorithmic framework for computing LLM chemistry, we now
ask a fundamental question: \emph{when does chemistry emerge in
model ensembles?} Answering this is critical for understanding the conditions
under which chemistry-based optimization is feasible and beneficial.

Our analysis reveals a surprising and counter-intuitive result: chemistry
emergence is intrinsically linked to model performance diversity, and perfect
models paradoxically eliminate the very interactions that chemistry-based
methods seek to exploit. This finding has profound implications for ensemble
design and optimization strategy selection.

We establish our main theoretical contributions through two key
results. First, we prove that chemistry emergence is equivalent to ensemble
performance heterogeneity, while homogeneous ensembles yield zero chemistry
regardless of individual model strength. Second, we show that chemistry
scales monotonically with performance diversity, though the optimal
level varies by task requirements. We now formalize our first result by proving
Theorem \ref{thm:chem}.

\begin{proof}
If all models share the same performance $(q_{i}^*, a_{i}^*)$ profile, then each has identical
penalty $p^* = (1 - q^{\mathit{norm}*}_i)(1 - a_{i}^*)$. By the linearity property,
$\mathit{cost}(X) = p^* \sum_{i=1}^k w_i$ depends only on set size, so
$\mathit{benefit}(\{a\}, X)$ depends only on $|X|$. Adding $b$ changes only the
set size, not the marginal effect of $a$. Thus $$\Delta(a,b,X) = 0 \quad
\Rightarrow \quad \mathit{chem}(a,b,S) = 0.$$ If models' performance differ, penalties vary, so
adding $a$ or $b$ changes costs in model-specific ways. Then
$\mathit{benefit}(\{a\}, X) \neq \mathit{benefit}(\{a\}, X \cup \{b\})$ for some
$X$, giving $\Delta(a,b,X) > 0$.
\end{proof}

We next prove Corollary \ref{cor:chem}, linking chemistry growth to model performance diversity.

\begin{proof}
By monotonicity, diverse performance $(q_i, a_i)$ profiles create diverse penalties, and by
submodularity, this diversity has diminishing returns. Greater performance diversity
increases variance in penalties, $\mathrm{Var}[\mathit{penalty}_i]$. This
amplifies selective pressure, producing more variation in benefits across
subsets. Since LLM chemistry depends on $$\max_X \frac{\Delta(a,b,X)}{\mathit{cost}(X \cup
\{a,b\})},$$ larger benefit differences yield larger $\Delta$, hence greater
chemistry. The precise ``beneficial'' diversity depends on task demands: high for
complementary reasoning, lower for factual consistency.
\end{proof}

\subsection{Representative Ensembles: Chemistry vs Performance Baseline}
\label{sec:ext-results}

Table~\ref{tab:best_ensembles} reports representative ensembles for each task
and group size, showing the model compositions and effectiveness of the best
observed LLM configurations (yielded by Algorithm~\ref{fig:recommend}) compared with the one yielded by the Performance baseline.
These examples complement the average results in Table~\ref{tab:rq1_performance}
by highlighting which specific model combinations achieved peak
effectiveness.

\begin{table}[htbp]
\centering
\scriptsize
\setlength{\tabcolsep}{4pt}
\renewcommand{\arraystretch}{1.15}
\caption{Best observed ensembles (Chemistry vs Best (Performance) baseline) for
Statement Classification, Automated Program Repair, and Clinical Note
Summarization. Each ensemble corresponds to the configuration achieving the
highest single-run effectiveness.}
\label{tab:best_ensembles}
\begin{adjustbox}{max width=\linewidth}
\begin{tabular}{ccp{0.28\linewidth}c p{0.28\linewidth}c}
\hline
\textbf{Task} & \textbf{Group Size} & \textbf{Chemistry (models)} & \textbf{Effectiveness} & \textbf{Best Baseline (PERF, models)} & \textbf{Effectiveness} \\
\hline
\multirow{3}{*}{\shortstack{Statement Credibility \\ Classification}} 
 & $3$ &
 \begin{itemize}[left=0pt,noitemsep,topsep=0pt,label={}]
   \item o4-mini
   \item mixtral:8x22b
   \item o1-mini
 \end{itemize} & 0.600 &
 \begin{itemize}[left=0pt,noitemsep,topsep=0pt,label={}]
   \item o4-mini
   \item claude-3-5-sonnet-latest
   \item o1-mini
 \end{itemize} & 0.600 \\
 & $5$ &
 \begin{itemize}[left=0pt,noitemsep,topsep=0pt,label={}]
   \item gpt-4o
   \item gemini-2.0-flash
   \item claude-3-5-sonnet-latest
   \item claude-3-7-sonnet-20250219
   \item firefunction-v2
 \end{itemize} & 0.700 &
 \begin{itemize}[left=0pt,noitemsep,topsep=0pt,label={}]
   \item claude-3-5-sonnet-latest
   \item claude-3-7-sonnet-20250219
   \item o4-mini
   \item qwen2.5:32b
 \end{itemize} & 0.700 \\
 & $10$ &
 \begin{itemize}[left=0pt,noitemsep,topsep=0pt,label={}]
   \item firefunction-v2
   \item claude-3-7-sonnet-20250219
   \item o1-mini
   \item gpt-4o
   \item gemini-2.0-flash
   \item claude-3-5-sonnet-latest
   \item qwen2.5:32b
   \item o4-mini
   \item o3-mini
 \end{itemize} & 0.800 &
 \begin{itemize}[left=0pt,noitemsep,topsep=0pt,label={}]
   \item o4-mini
   \item gemini-2.0-flash
   \item o3-mini
   \item gpt-4o
   \item claude-3-7-sonnet-20250219
   \item o1-mini
   \item claude-3-5-sonnet-latest
   \item firefunction-v2
   \item qwen2.5:32b
 \end{itemize} & 0.800 \\
\hline
\multirow{3}{*}{\shortstack{Clinical Notes \\Summarization}} 
 & $3$ &
 \begin{itemize}[left=0pt,noitemsep,topsep=0pt,label={}]
   \item firefunction-v2
   \item gemini-2.0-flash
   \item claude-3-7-sonnet-20250219
 \end{itemize} & 0.700 &
 \begin{itemize}[left=0pt,noitemsep,topsep=0pt,label={}]
   \item mixtral:8x22b
   \item gpt-4o
   \item qwen2.5:32b
 \end{itemize} & 1.000 \\
 & $5$ &
 \begin{itemize}[left=0pt,noitemsep,topsep=0pt,label={}]
   \item firefunction-v2
   \item gemini-2.0-flash
   \item qwen2.5:32b
   \item gpt-4o
   \item mixtral:8x22b
 \end{itemize} & 1.000 &
 \begin{itemize}[left=0pt,noitemsep,topsep=0pt,label={}]
   \item gpt-4o
   \item o3-mini
   \item llama3.1:70b
   \item mixtral:8x22b
   \item qwen2.5:32b
 \end{itemize} & 1.000 \\
 & $10$ &
 \begin{itemize}[left=0pt,noitemsep,topsep=0pt,label={}]
   \item mixtral:8x22b
   \item firefunction-v2
   \item gpt-4o
   \item o4-mini
   \item o3-mini
   \item gemini-2.0-flash
   \item claude-3-5-sonnet-latest
   \item llama3.3:latest
   \item o1-mini
   \item qwen2.5:32b
 \end{itemize} & 1.000 &
 \begin{itemize}[left=0pt,noitemsep,topsep=0pt,label={}]
   \item gemini-2.0-flash
   \item o3-mini
   \item mixtral:8x22b
   \item o1-mini
   \item llama3.1:70b
   \item firefunction-v2
   \item gpt-4o
   \item o4-mini
   \item llama3.3:latest
   \item qwen2.5:32b
 \end{itemize} & 1.000 \\
\hline
\multirow{3}{*}{\shortstack{Automated Program \\Repair}} 
 & $3$ &
 \begin{itemize}[left=0pt,noitemsep,topsep=0pt,label={}]
   \item firefunction-v2
   \item claude-3-7-sonnet-20250219
   \item claude-3-5-sonnet-latest
 \end{itemize} & 1.000 &
 \begin{itemize}[left=0pt,noitemsep,topsep=0pt,label={}]
   \item gpt-4o
   \item claude-3-5-sonnet-latest
   \item o4-mini
 \end{itemize} & 1.000 \\
 & $5$ &
 \begin{itemize}[left=0pt,noitemsep,topsep=0pt,label={}]
   \item o3-mini
   \item claude-3-7-sonnet-20250219
   \item gpt-4o
   \item mixtral:8x22b
   \item llama3.1:70b
 \end{itemize} & 1.000 &
 \begin{itemize}[left=0pt,noitemsep,topsep=0pt,label={}]
   \item o1-mini
   \item firefunction-v2
   \item gpt-4o
   \item claude-3-5-sonnet-latest
   \item o4-mini
 \end{itemize} & 1.000 \\
 & $10$ &
 \begin{itemize}[left=0pt,noitemsep,topsep=0pt,label={}]
   \item o1-mini
   \item firefunction-v2
   \item o3-mini
   \item claude-3-7-sonnet-20250219
   \item qwen2.5:32b
   \item gpt-4o
   \item claude-3-5-sonnet-latest
   \item mixtral:8x22b
   \item llama3.1:70b
   \item o4-mini
 \end{itemize} & 1.000 &
 \begin{itemize}[left=0pt,noitemsep,topsep=0pt,label={}]
   \item o1-mini
   \item firefunction-v2
   \item o3-mini
   \item claude-3-7-sonnet-20250219
   \item qwen2.5:32b
   \item gpt-4o
   \item claude-3-5-sonnet-latest
   \item mixtral:8x22b
   \item o4-mini
   \item llama3.3:latest
 \end{itemize} & 1.000 \\
\hline
\end{tabular}
\end{adjustbox}
\begin{minipage}{1.\linewidth}
\scriptsize
\textbf{Note:} For Clinical Note Summarization (group size 3), Chemistry
dropped ($0.700$ vs. $1.000$) as the \texttt{claude-3-7-sonnet-20250219} model underperformed. This is related
to temporary service issues or budget limits at the time of evaluation.
\end{minipage}
\end{table}

\subsection{Interpretive Analysis of Chemistry Emergence}
\label{sec:interpretation}

Having established Theorem \ref{thm:chem} and Corollary \ref{cor:chem}, we now
provide a visual interpretation through chemistry maps, which illustrate how the
predicted conditions for LLM chemistry emergence manifest across the
accuracy-quality space. These maps display the marginal complementarity ($\Delta
\mathrm{CI}$) of adding a new model to an ensemble, serving as a visual proxy
for the theorem's guarantees. In particular, our theorem and corollary link
chemistry to model performance heterogeneity: homogeneity implies $\Delta = 0$ and thus
$\mathit{chem}=0$, while heterogeneity implies $\Delta > 0$ and thus
$\mathit{chem}>0$. The maps make this relationship visible—bright regions mark
conditions under which chemistry would emerge ($\Delta > 0$, heterogeneity),
whereas dark regions reflect redundancy ($\Delta = 0$, homogeneity).
Intuitively, ensembles with many bright regions have substantial room for
chemistry to emerge and improve performance, ensembles with only tiny bright
slivers allow for marginal gains at best, and fully dark maps indicate
\textit{saturation}, where new members cannot increase CI and performance gains
from chemistry are no longer possible. In rare cases, the maps appear nearly
uniform with negligible variation in $\Delta \mathrm{CI}$; these are interpreted
as \textit{saturated ensembles} and not as evidence of broad chemistry. We
present chemistry maps for three tasks of varying complexity:

\begin{figure*}[t]
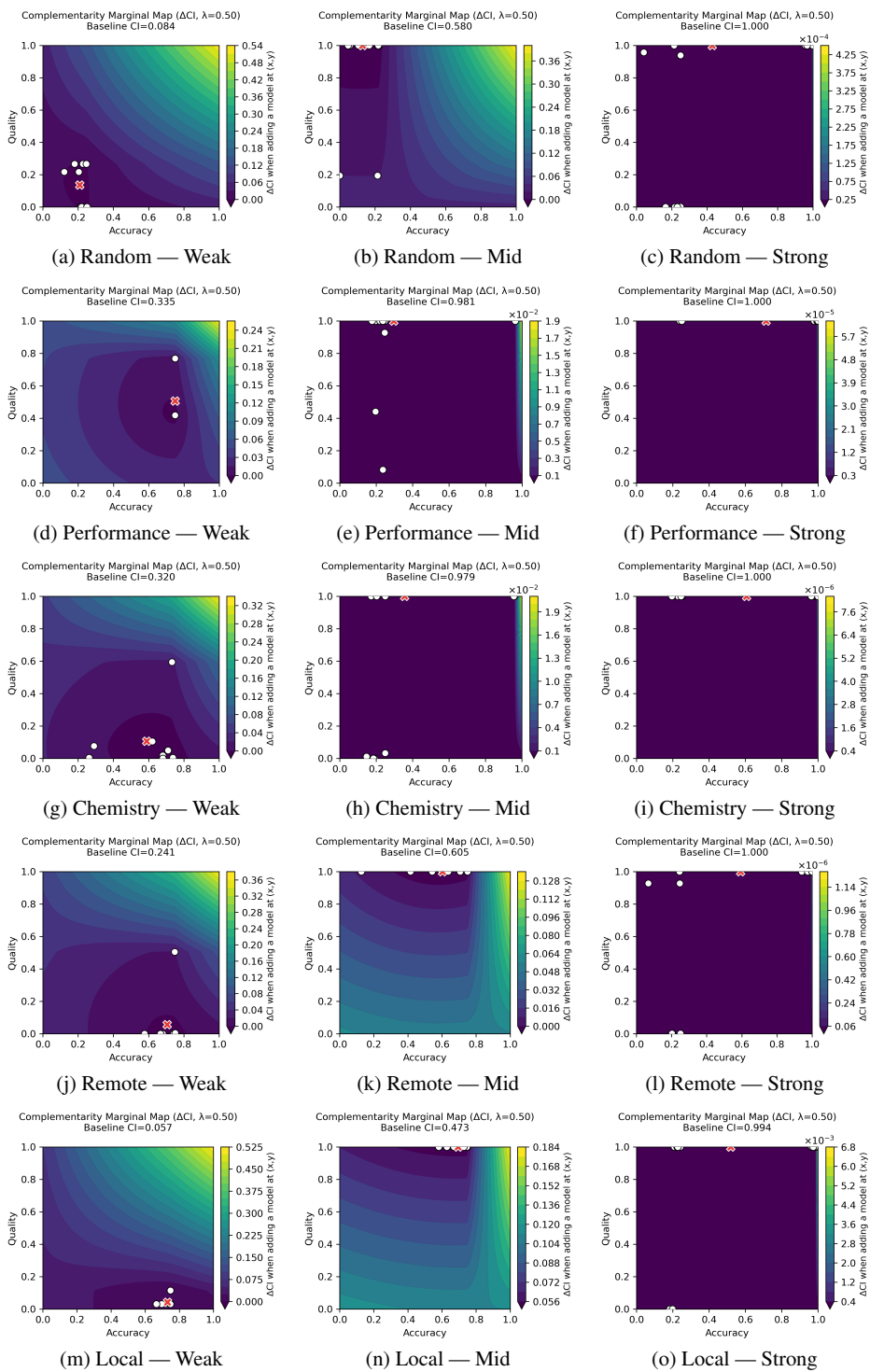

  \centering
  \scalebox{0.95}{
    \begin{minipage}{\textwidth}
      \chemrow{task_A}{group3_random}{Random}
      \chemrow{task_A}{group3_performance}{Performance}
      \chemrow{task_A}{group3_chemistry}{Chemistry}
      \chemrow{task_A}{group3_remote}{Remote}
      \chemrow{task_A}{group3_local}{Local}
    \end{minipage}
  }

  \caption{LLM chemistry maps (marginal complementarity, \(\Delta\mathrm{CI}\),
  trade-off parameter \(\lambda=0.5\)) for \textbf{Statement Credibility
  Classification} (\textit{low complexity}, $N = 10$). Rows correspond to
  strategies (Random, Performance, Remote, Local); the Chemistry row is included
  for comparison. Columns show \textbf{Weak}, \textbf{Mid}, and \textbf{Strong}
  ensembles. Weak ensembles display extensive bright regions, indicating
  substantial chemistry potential and performance gains. Mid ensembles are mixed,
  with some retaining bright regions and others already saturated. Strong
  ensembles are almost entirely dark, reflecting saturation where added models are
  redundant. A few weak panels appear nearly uniform in $\Delta \mathrm{CI}$;
  these reflect negligible variation and are treated as saturated rather than as
  broad chemistry.}
  \label{fig:chemistry-grid-taskA}

\end{figure*}

\begin{figure*}[t]
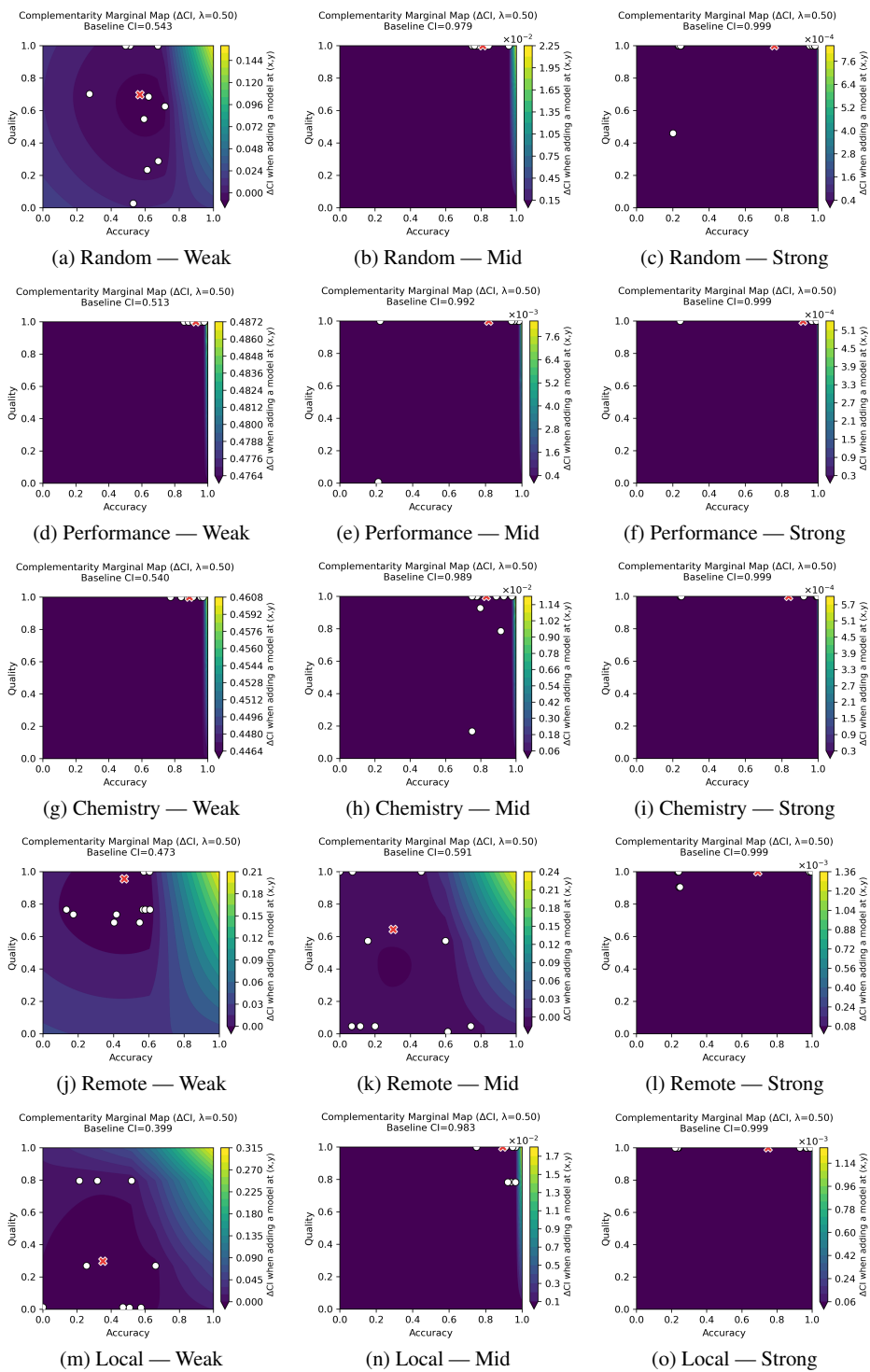

  \centering
  \scalebox{0.95}{
    \begin{minipage}{\textwidth}
      \chemrow{task_C}{group3_random}{Random}
      \chemrow{task_C}{group3_performance}{Performance}
      \chemrow{task_C}{group3_chemistry}{Chemistry}
      \chemrow{task_C}{group3_remote}{Remote}
      \chemrow{task_C}{group3_local}{Local}
    \end{minipage}
  }

  \caption{LLM chemistry maps (marginal complementarity, \(\Delta\mathrm{CI}\),
  trade-off parameter \(\lambda=0.5\)) for \textbf{Clinical Notes Summarization}
  (\textit{medium complexity}, $N = 10$). Rows correspond to strategies
  (Random, Performance, Remote, Local); the Chemistry row is included for
  comparison. Columns show \textbf{Weak}, \textbf{Mid}, and \textbf{Strong}
  ensembles. Weak ensembles are mixed: three maps display bright regions
  (chemistry emergence possible), while two are mostly dark with marginal
  potential. Mid ensembles are largely dark, with one strategy retaining notable
  bright regions but most showing only small gains. Strong ensembles are uniformly
  dark with tiny slivers, indicating near-complete saturation where added models
  provide little benefit.}
  \label{fig:chemistry-grid-taskC}

\end{figure*}

\begin{figure*}[t]
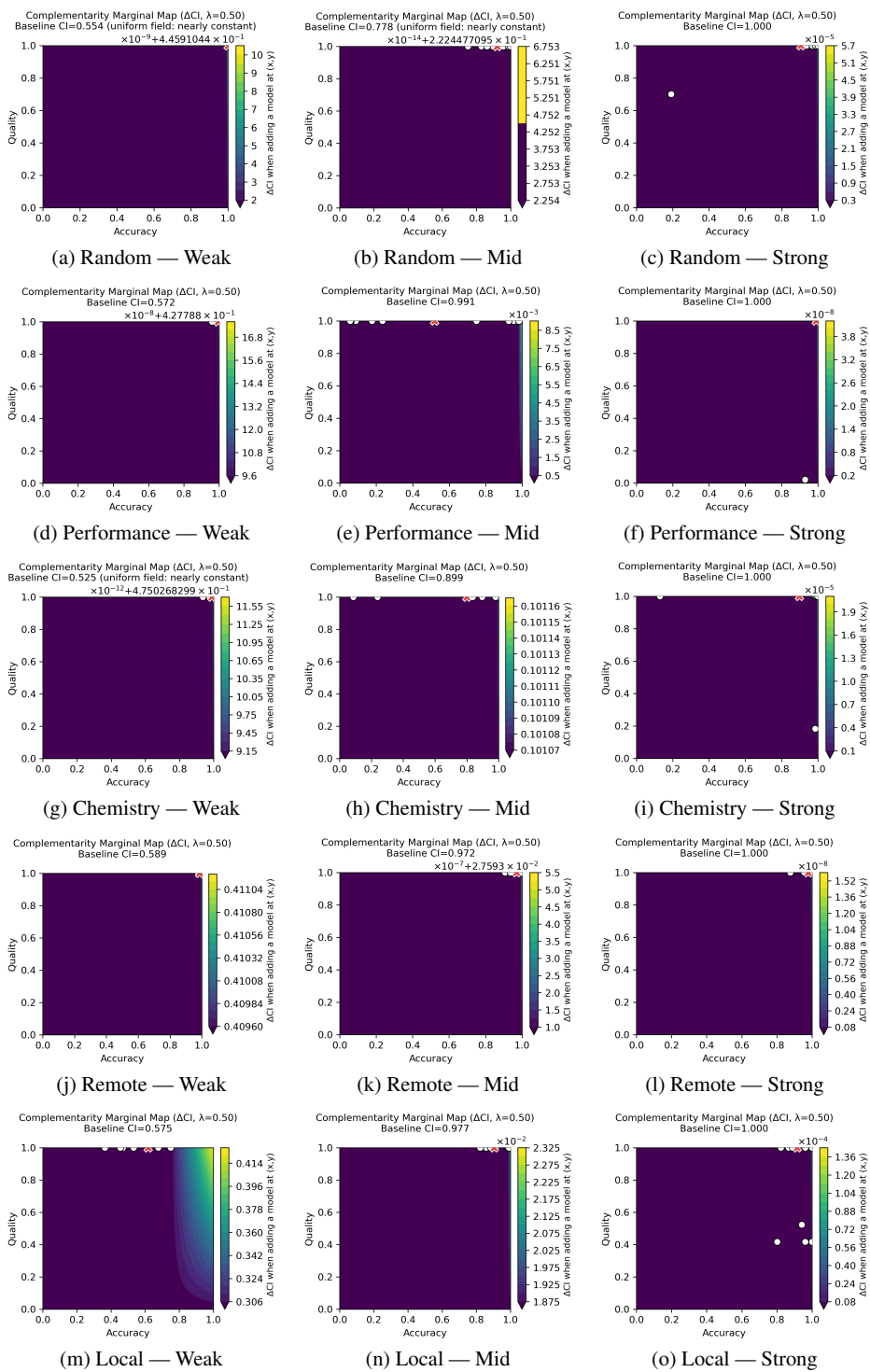

  \centering
  \scalebox{0.95}{
    \begin{minipage}{\textwidth}
    \chemrow{task_B}{group3_random}{Random}
    \chemrow{task_B}{group3_performance}{Performance}
    \chemrow{task_B}{group3_chemistry}{Chemistry}
    \chemrow{task_B}{group3_remote}{Remote}
    \chemrow{task_B}{group3_local}{Local}
    \end{minipage}
  }
  \caption{LLM chemistry maps (marginal complementarity, \(\Delta\mathrm{CI}\),
  trade-off parameter \(\lambda=0.5\)) for \textbf{Automated Program Repair}
  (\textit{high complexity}, $N = 10$). Rows correspond to strategies
  (Random, Performance, Remote, Local); the Chemistry row is included for
  comparison. Columns show \textbf{Weak}, \textbf{Mid}, and \textbf{Strong}
  ensembles. Bright regions are nearly absent across all ensembles, indicating
  saturation (\(\Delta\mathrm{CI} \approx 0\)) where added models are redundant,
  chemistry emergence is negligible, and performance plateaus. Occasional uniform
  panels with vanishing $\Delta \mathrm{CI}$ variation are likewise interpreted as
  saturated.}
  \label{fig:chemistry-grid-taskB}
\end{figure*}

\subsection{Benchmark Sample Runs}
\label{ssec:trial-runs}

Each benchmark run produces a \texttt{CSV} file that records the outputs and 
metrics of LLM executions. The file contains one row for each 
$(\textit{trial}, \textit{model}, \textit{task})$ combination, together with 
the performance and metadata fields listed in 
Table~\ref{tab:trial-columns}. A collection of such files constitutes our 
\emph{performance histories} (See Figure~\ref{fig:overview-chemistry}'s first step).

\begin{table}[htb]
\centering
\caption{Columns present in the CSV file produced by benchmark runs.}
\label{tab:trial-columns}
\small
\begin{tabular}{lcl}
\hline
\multicolumn{1}{l}{\textbf{Column}} & \multicolumn{1}{l}{\textbf{Data Type}} & \multicolumn{1}{c}{\textbf{Description}} \\ \hline
trial & string & Name or identifier of the benchmark run. \\ \hline
model & string & Name of the evaluated LLM. \\ \hline
task & string & Input task text (prompt or query $Q$). \\ \hline
latency & float & Time in seconds to generate the LLM output. \\ \hline
temperature & float & Sampling temperature used for generation. \\ \hline
id & string & Unique identifier for the generated output. \\ \hline
result & string & Raw text output produced by the LLM. \\ \hline
quality & float & Consensus-based quality score of the generated output. \\ \hline
gen\_accuracy & float & Generation accuracy of the initial generation step. \\ \hline
variance & float & Variance of quality scores (dispersion/confidence measure). \\ \hline
review\_accuracy & float & Review accuracy estimated via MVLE (inverse of variance). \\ \hline
accuracy & float & Final accuracy score, combining generation and review. \\ \hline
elapsed & string & Human-readable elapsed generation time. \\ \hline
created & string & Human-readable timestamp of output creation. \\ \hline
\end{tabular}
\end{table}

\newpage

\begin{tcolorbox}[fontupper=\small, fontlower=\footnotesize,title=\textbf{Task:}~\emph{Classify a short statement into a category of fakeness},boxrule=1pt, sharp corners,colback=white]
    \begin{tcolorbox}[fontupper=\footnotesize, fontlower=\scriptsize,title=\textbf{model:} gpt-4o,colback=white]
        \textbf{result:} "The Chicago Bears have had more starting quarterbacks in the last 10 years than the total number of tenured (UW) faculty fired during the last two decades." is labeled as "true".\\
        \textbf{latency:} $2.0$\\
        \textbf{quality:} $1\times 10^{-10}$\\
        \textbf{gen\_accuracy:} $0.0$\\
        \textbf{review\_accuracy:} $0.9999999998482016$\\
        \textbf{accuracy}: $0.2499999999620504$
    \end{tcolorbox}

    \begin{tcolorbox}[fontupper=\footnotesize, fontlower=\scriptsize,title=\textbf{model:} o3-mini,colback=white]
        \textbf{result:} "'I'm the only person on this stage who has worked actively just last year passing, along with Russ Feingold, some of the toughest ethics reform since Watergate.' is labeled as 'half-true'."\\
        \textbf{latency:} $31.0$\\
        \textbf{quality:} $7.1397913333$\\
        \textbf{gen\_accuracy:} $1.0$\\
        \textbf{review\_accuracy:} $0.8969102035156231$\\
        \textbf{accuracy}: $0.9742275508789058$
    \end{tcolorbox}

    \begin{tcolorbox}[fontupper=\footnotesize, fontlower=\scriptsize,title=\textbf{model:} llama3.1:70b,colback=white]
        \textbf{result:} "'However, it took \$19.5 million in Oregon Lottery funds for the Port of Newport to eventually land the new NOAA Marine Operations Center-Pacific.' is labeled as 'true'."\\
        \textbf{latency:} $82.0$\\
        \textbf{quality:} $1.9353739253$\\
        \textbf{gen\_accuracy:} $0.0$\\
        \textbf{review\_accuracy:} $0.224749679737892$\\
        \textbf{accuracy}: $0.056187419934473$
    \end{tcolorbox}

    \begin{tcolorbox}[fontupper=\footnotesize, fontlower=\scriptsize,title=\textbf{model:} gemini-2.0-flash,colback=white]
        \textbf{result:} "'The economic turnaround started at the end of my term.' is labeled as 'half-true'."\\
        \textbf{latency:} $2.0$\\
        \textbf{quality:} $9.0630267113$\\
        \textbf{gen\_accuracy:} $1.0$\\
        \textbf{review\_accuracy:} $0.8617875069130844$\\
        \textbf{accuracy}: $0.9654468767282711$
    \end{tcolorbox}
\end{tcolorbox}

\end{document}